\documentclass[11pt]{article}

\usepackage[preprint]{acl}
\usepackage{multirow}
\usepackage{booktabs}
\usepackage{times}
\usepackage{latexsym}
\usepackage{hyperref}
\usepackage[T1]{fontenc}
\usepackage{multirow}
\usepackage[utf8]{inputenc}

\usepackage{caption}
\usepackage{microtype}

\usepackage{inconsolata}

\usepackage{graphicx}

\usepackage[most]{tcolorbox}
\usepackage{svg}
\usepackage{amsmath}
\usepackage{algorithm,algorithmic}
\usepackage{listings}
\tcbset{
  myhighlight/.style={
    colback=gray!20,    
    colframe=gray!40,   
    boxrule=0pt,        
    arc=2mm,            
    boxsep=0pt,         
    enhanced,
    sharp corners=downhill, 
  }
}

\title{LightAutoDS-Tab: Multi-AutoML Agentic System for Tabular Data}

\author{
 \textbf{Aleksey Lapin\textsuperscript{2}}, 
 \textbf{Igor Hromov\textsuperscript{1}}, 
 \textbf{Stanislav Chumakov\textsuperscript{2}},
 \textbf{Mile Mitrovic\textsuperscript{1}},
 \textbf{Dmitry Simakov\textsuperscript{1}},\\
 \textbf{Nikolay O. Nikitin\textsuperscript{2}},
 \textbf{Andrey V. Savchenko\textsuperscript{1}}
\\
 \textsuperscript{1} Sber AI Lab,
 \textsuperscript{2} ITMO University, 
\\
}

\begin{document}
\maketitle
\begin{abstract}



AutoML has advanced in handling complex tasks using the integration of LLMs, yet its efficiency remains limited by dependence on specific underlying tools. In this paper, we introduce LightAutoDS-Tab, a multi-AutoML agentic system for tasks with tabular data, which combines an LLM-based code generation with several AutoML tools. Our approach improves the flexibility and robustness of pipeline design, outperforming state-of-the-art open-source solutions on several data science tasks from Kaggle. The code of LightAutoDS-Tab is available in the open repository \url{https://github.com/sb-ai-lab/LADS}

\end{abstract}

\section{Introduction}
\label{sec_intro}

Classic tabular AutoML tools, such as AutoGluon \cite{erickson2020autogluontabularrobustaccurateautoml} and H2O \cite{ledell2020h2o} generally rely on predefined search spaces and routines, often focusing on hyperparameter optimization and model ensembling. Leveraging natural language understanding and code generation abilities of large language models (LLMs) has recently led to the emergence of LLM-based agents capable of automating parts of the ML workflow. General strategies involve planning and iterative plan refinement \cite{guo2024dsagentautomateddatascience}, a tree-based search process with agentic trial-and-error code construction \cite{jiang2025aideaidrivenexplorationspace}, and fully autonomous agentic systems \cite{wang2024openhandsopenplatformai} set for ML (Machine Learning) problems. 

While all of the mentioned strategies result in a robust search space across preliminary model training stages and broad flexibility of the pipeline design, they are prone to excessive branching before reaching an optimal solution, where each search branch must include task descriptions, historical experience (demonstrations), knowledge, intermediate reasoning steps, generated code, error logs, and feedback. Appending conversation history or providing detailed context in prompts can face limitations due to the limited LLM context window, drastically increasing cost while not guaranteeing model composition performance on the latter pipeline stages, which is better handled by classic AutoML frameworks. LLM-based solutions for AutoML tasks are implemented in
AutoKaggle \cite{li2024autokagglemultiagentframeworkautonomous}, a multi-agentic approach for model selection, and AIDE (AI-Driven Exploration) \cite{jiang2025aideaidrivenexplorationspace}, which performs a tree search over solutions. However, such techniques do not use the potential of various existing ML and AutoML tools. Other existing solutions are joined with one specific tool, e.g., AutoGluon in AutoGluon Assistant~\footnote{\url{https://github.com/autogluon/autogluon-assistant}}~\cite{fang2025mlzero}, and inherit their disadvantages.

\begin{figure*}[t]
\centerline{
\includegraphics[width=1.0\textwidth]{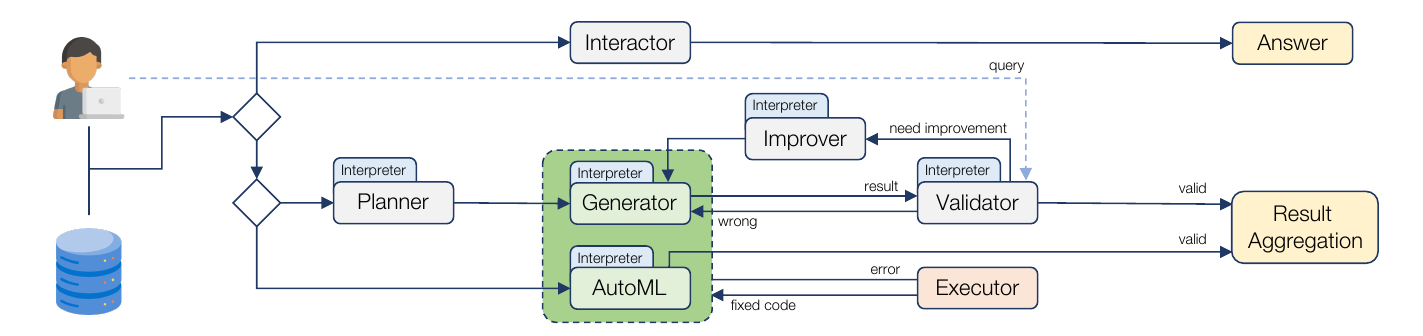}
}
\caption{The proposed LightAutoDS-Tab multi-AutoML agentic system architecture.}
\label{fig_schema}
\end{figure*}

To address these limitations, we introduce LightAutoDS-Tab (Figure~\ref{fig_schema}), a novel multi-AutoML framework that enhances the capabilities of various existing AutoML tools by integrating them with an LLM agent. The latter performs flexible, data-aware code generation for early-stage tasks of the ML pipeline and configuration of the dedicated AutoML framework. It allows for achieving effective and robust pipeline construction, thereby addressing the limitations of LLM-based and traditional fixed-pipeline AutoML techniques. The code is generated around a skeleton that covers all the essential steps of an ML pipeline, adding operations that increase search effectiveness. The stages include task understanding, data preprocessing steps based on exploratory data analysis, configuring a dedicated AutoML framework, and evaluating and debugging. The ease of use, low operational cost, and export of completed scripts make LightAutoDS-Tab suitable for fast prototyping by experienced data scientists and an approachable learning and quick-start tool for beginners in ML. We experimentally demonstrate that our LightAutoDS-Tab improves quality and efficiency compared to the known state-of-the-art techniques~\citep{jiang2025aideaidrivenexplorationspace,li2024autokagglemultiagentframeworkautonomous} on classic and recent Kaggle competitions, simulating real-world data science problems. 

\section{Related works}


\subsection{AutoML tools}

The problem of automating ML processes and improving accessibility has been actively researched in recent years, leading to significant advances in AutoML. Numerous AutoML tools have been developed and implemented, such as AutoGluon \cite{erickson2020autogluontabularrobustaccurateautoml}, H2O AutoML \cite{ledell2020h2o}, LightAutoML \cite{vakhrushevlightautoml}, and FEDOT \cite{nikitin2022automated}. AutoML tools help accelerate development and improve model quality, yet their use remains the prerogative of experienced data scientists, primarily due to the complicated configuration process.

\subsection{Data Science agents}

Various agents are aimed at solving DS tasks in a more automated way. In MLCopilot \cite{zhang2024mlcopilotunleashingpowerlarge}, knowledge-based reasoning is proposed as a form of effective utilization of historical data. The workflow starts with collecting an experience pool, which is then used in an online stage to retrieve relevant examples for the current task and add them to the experience pool. The method addresses tabular, textual, and image data. 

In AutoKaggle \cite{li2024autokagglemultiagentframeworkautonomous}, the multi-agentic approach involves decomposing the ML workflow into distinct phases, such as data understanding, model prediction, and result summarization, each managed by specialized agents. Iterative debugging methods ensure correctness and consistency, while data cleaning and feature engineering are performed with the integrated tools library. However, the ML aspect is performed using a simple choice between regression, random forest, and XGBoost. 

In TabPFN \cite{hollmann2023tabpfntransformersolvessmall}, effective tabular classification is achieved with a transformer model pretrained on tabular model evaluations \cite{salinas2024tabrepolargescalerepository} and in-context learning (ICL) to approximate Bayesian inference.

In Tree-Search Enhanced LLM Agents (SELA) \cite{chi2024selatreesearchenhancedllm}, Monte Carlo Tree Search (MCTS) is used to expand the search space. The approach is enhanced in the exploration and exploitation phases in I-MCTS\cite{liang2025imctsenhancingagenticautoml} by adding introspective nodes and a hybrid reward system, combining LLM-driven value estimations with real-world performance feedback. 

Also, some agents are more specialized. For example, the authors of CAAFE \cite{hollmann2023largelanguagemodelsautomated} have designed a complete pipeline for feature generation using LLMs, which automatically generate and execute code for new features. 


Autonomous general-purpose agentic systems are another field of active growth with developments in workflow construction \cite{zhang2025aflowautomatingagenticworkflow} and success in autonomous coding benchmarks of general \cite{jimenez2024swebenchlanguagemodelsresolve} and data science-oriented \cite{huang2024dacodeagentdatascience} purpose. The authors of MLE-bench \cite{chan2024mle} test fully autonomous systems, which the authors refer to as scaffolds, such as AIDE \cite{jiang2025aideaidrivenexplorationspace}, MLAB \cite{huang2024mlagentbenchevaluatinglanguageagents} and CodeActAgent from the OpenHands platform \cite{wang2024openhandsopenplatformai}. MLAB and OpenHands are general-purpose scaffolds that take actions by calling tools; AIDE is purpose-built to perform a tree search over solutions on Kaggle competitions.

\section{Proposed Framework}
In this paper, we introduce LightAutoDS-Tab (Fig.~\ref{fig_schema}), an automated framework built on an agentic LLM workflow that streamlines the development of pipelines for tabular data and configuration setup for dedicated standard ML and AutoML tools.

\subsection{System Overview}\label{sub_sec_design}
\begin{algorithm}[ht]
\caption{LightAutoDS-Tab Agentic Workflow}
\begin{algorithmic}[1]
\renewcommand{\algorithmicrequire}{\textbf{Input:}}
\renewcommand{\algorithmicensure}{\textbf{Output:}}
\REQUIRE User query $\mathcal{Q}$, Dataset $\mathcal{D}$, LLM $\mathcal{L}$
\ENSURE Predictions $\mathcal{P}$, Code $\mathcal{C}$, Report $\mathcal{R}$, Answer $\mathcal{O}$
\STATE \textbf{// System Initialization}
\STATE $\mathcal{S} \gets \{\mathcal{M} \gets [\mathcal{Q}]\}$ \COMMENT{State: Messages $\mathcal{M}$}
\STATE $\mathcal{A}_{\text{int-act}} \gets$ \textsc{InteractorAgent}($\mathcal{S}, \mathcal{L}$)
\STATE $\mathcal{A}_{\text{plan}} \gets$ \textsc{PlannerAgent}($\mathcal{S}, \mathcal{L}, \mathcal{D}$)
\STATE $\mathcal{A}_{\text{gen}} \gets$ \textsc{GeneratorAgent}($\mathcal{S}, \mathcal{L},  \mathcal{D}$)
\STATE $\mathcal{A}_{\text{val}} \gets$ \textsc{ValidatorAgent}($\mathcal{S}, \mathcal{L},  \mathcal{D}$)
\STATE $\mathcal{A}_{\text{impr}} \gets$ \textsc{ImproverAgent}($\mathcal{S}, \mathcal{L},  \mathcal{D}$)
\STATE $\mathcal{A}_{\text{auto}} \gets$ \textsc{AutomlAgent}($\mathcal{S}, \mathcal{L}, \mathcal{D}$)
\STATE $\mathcal{A}_{\text{int-pret}} \gets$ \textsc{InterpreterAgent}($\mathcal{S}, \mathcal{L}$)
\STATE \textbf{// LightAutoDS-Tab Workflow Loop}
\REPEAT
    \STATE $\alpha_{\text{next}} \gets \textsc{UNDEFINED}$
    \STATE $\alpha_{\text{next}} \gets \mathcal{A}_{\text{sup}} \triangleright \textsc{SelectAgent}(\mathcal{S})$
    \IF{$\alpha_{\text{next}} = \mathcal{A}_{\text{int-act}}$}
        \RETURN $\mathcal{O}$
    \ENDIF
    \IF{$\alpha_{\text{next}} = \mathcal{A}_{\text{planner}}$}
        \REPEAT
        \STATE $\mathcal{A}_{\text{gen}} \gets$ $\mathcal{A}_{\text{plan}}$
        \STATE $\mathcal{A}_{\text{vla}} \gets$ $\mathcal{A}_{\text{gen}}$
        \STATE $\mathcal{A}_{\text{impr}} \gets$ $\mathcal{A}_{\text{vla}}$
        \STATE $\mathcal{A}_{\text{gen}} \gets$ $\mathcal{A}_{\text{impr}}$
        \UNTIL{$\mathcal{A}_{\text{val}} = \textsc{valid}$} // \textcolor{teal}{Goal satisfied}
        \RETURN $\mathcal{P}$, $\mathcal{C}$, $\mathcal{R}$
    \ENDIF
    \IF{$\alpha_{\text{next}} = \mathcal{A}_{\text{auto}}$}
        \RETURN $\mathcal{P}$, $\mathcal{C}$, $\mathcal{R}$
    \ENDIF
\UNTIL {$\mathcal{S} = \emptyset$} // \textcolor{teal}{Chat ended}
\end{algorithmic}
\label{alg_1}
\end{algorithm}

\begin{figure*}[t]
\centerline{
\includegraphics[width=1.\textwidth]{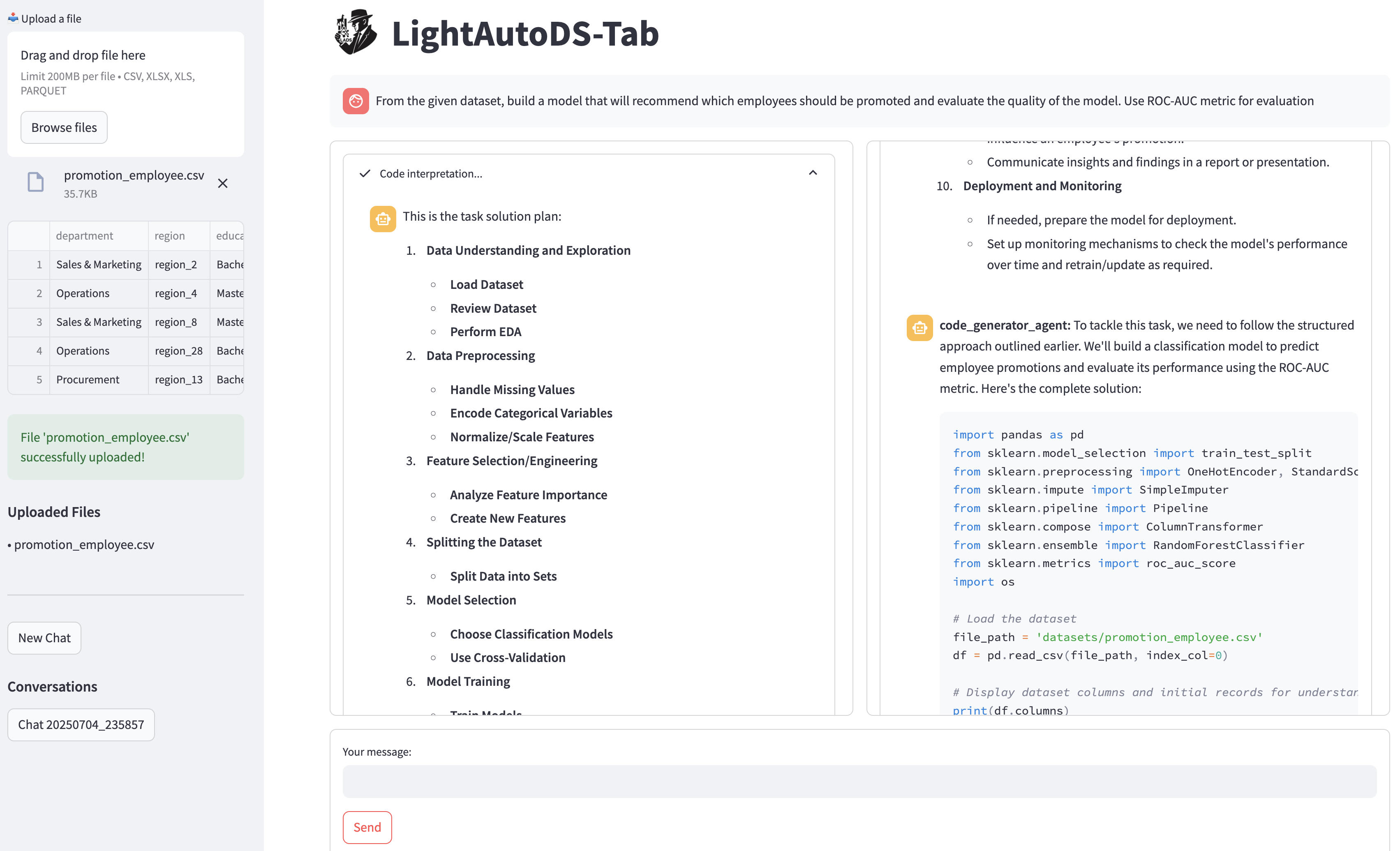}
}
\caption{The LightAutoDS-Tab user interface.}
\label{fig_ui}
\end{figure*}

LightAutoDS-Tab presents a multi-AutoML agentic system that combines the strengths of LLMs and the existing ML and AutoML tools. The architecture of the system is illustrated in Figure~\ref{fig_schema}.

The system includes several specialized agents responsible for specific tasks such as code generation and debugging, result validation, explanation generation (technical and non-technical), tool invocation, and user interaction. Users only need to provide the dataset and a query outlining their desired objective. Upon receiving this input, the system clarifies the user’s intent and determines whether the query relates to building an ML pipeline or requires direct user interaction.

If the query involves user interaction, the \textbf{Interactor} agent provides relevant explanations on the system processes to ensure user understanding. When the query requests building an ML pipeline, the system activates an additional routing mechanism. This router evaluates the input and determines the appropriate pathway, allowing the system to proceed along one of two distinct routes:
\begin{enumerate}
  \item \textbf{LLM-Driven Pipeline Generation}:
  In this route, the Planner agent first formulates a plan for creating the ML pipeline based on the task description. The \textbf{Generator} agent, powered solely by LLMs, generates code using standard ML libraries such as Scikit-learn, CatBoost, TabPFN, and others. The generated code is subsequently evaluated by the \textbf{Validator} agent to ensure correctness and performance. If the validation criteria are not met, the \textbf{Improver} agent iteratively refines the code to enhance its performance metrics until satisfactory results are achieved.
  \item \textbf{AutoML-Based Pipeline Configuration}:
  Alternatively, the AutoML agent invokes tools such as LightAutoML or FEDOT, where key parameters are extracted from the task description using LLM for framework configurations.  
\end{enumerate}

We include several AutoML tools that are efficient in various tasks. For example, LightAutoML emerged as the winner of Kaggle AutoML Grand Prix 2024 \footnote{\url{https://www.kaggle.com/automl-grand-prix}} — the only solution that secured first place in two out of the five tasks. Participants competed in solving five machine learning tasks, with only 24 hours allotted for each task and the freedom to use various AutoML systems. As an additional AutoML tool, FEDOT is used since it allows the building of flexible pipelines for complicated tasks.

The system can either automatically choose the most appropriate route based on the characteristics of the input task or it can follow explicit instructions provided by the user in the query. This flexible routing ensures that both customization and automation are available when needed.

Code execution and debugging are handled by the \textbf{Executor} tool. Throughout the process, the \textbf{Interpreter} agent provides non-technical reports of each step, enhancing interpretability. These reports are compiled and summarized into a comprehensive final report, ensuring transparency and traceability of the pipeline design process, along with the corresponding code and trained model.

The end-to-end LightAutoDS-Tab agentic workflow is shown in Algorithm~\ref{alg_1}.

\subsection{Multi-AutoML Agentic Implementation}

As mentioned in Subsection~\ref{sub_sec_design}, standard ML frameworks are typically accessed through LLMs by generating end-to-end ML pipelines. At the same time, AutoML tools are treated as pre-packaged solutions where the LLM tunes key parameters. However, one major challenge preventing the generation of effective AutoML solutions is that even advanced reasoning and code-specialized LLMs can not produce functional code consistently, even after multiple refinement cycles. Several attempts have been made to enhance this process using retrieval-augmented generation (RAG) techniques, leveraging code repositories and documentation, but these efforts have not yet yielded successful results.

This limitation highlights the need for a new research direction focused on developing AutoML frameworks through agent-based approaches. Although the current LightAutoDS-Tab system may not yet be capable of directly improving AutoML model accuracy, it significantly enhances automation, reducing both the time and effort required to build robust, error-free AutoML pipelines.

The examples of prompts are provided in Appendix~\ref{app_a}.

\subsection{User interface}
The LightAutoDS-Tab framework features a user-friendly interface (UI)  developed with Streamlit (Figure~\ref{fig_ui}) and is designed to streamline the interaction between users and the automated data science workflow. The UI enables users to upload and preview datasets in standard formats such as \textit{csv}, \textit{xlsx}, and \textit{parquet}.

Users define their task once a dataset is uploaded by entering a query. This query can be written in varying levels of detail: advanced users with data science expertise can provide technical instructions. At the same time, business-oriented descriptions allow non-experts to utilize the platform effectively.

The interface includes two main panels that display how LightAutoDS-Tab processes and solves the given task in real-time, significantly enhancing interpretability:

\begin{enumerate}
  \item The \textit{right panel} provides detailed technical insights into each step of the ML pipeline construction, offering transparency for expert users.
  \item The \textit{left panel} presents a simplified, non-technical summary of the process, making it easy for non-experts to follow along and understand the results.
\end{enumerate}
Upon completing the task, the UI allows users to generate benchmark tables comparing performance metrics across different models or agent runs. These results can be manually added via a \textit{benchmark\_results.csv} file for comparative analysis. 

Furthermore, LightAutoDS-Tab includes a dedicated inference module that enables efficient integration into production environments. This module automates the generation of deployment-ready code alongside trained models, enabling them to be automatically invoked for standalone prediction tasks.

LightAutoDS-Tab can be easily extended to other tools like AutoGluon. It integrates with multiple LLM providers, and the system architecture allows for the seamless addition of new LLM-based providers when needed.   


\subsection{Main advantages}
LightAutoDS-Tab is designed to boost the productivity of Data Scientists and Analysts working with tabular data by focusing on the following key objectives:

\textbf{Increase Automation:} Streamline the development and optimization of ML pipelines, using natural language for intuitive instruction and configuration.

\textbf{Reduce Development Time:} Minimize the effort and time required to build end-to-end ML pipelines, accelerating the path from idea to deployable code.

\textbf{Maintain Interpretability:} Ensure transparency in pipeline construction with clear explanations accessible from technical and non-technical perspectives.

\textbf{Improve Quality:} Build and train models that deliver strong and reliable performance in validation metrics.

\section{Usage and evaluation}

\subsection{Cases and benchmarks}

In the experimental part, we follow the authors of AutoKaggle \cite{li2024autokagglemultiagentframeworkautonomous} to benchmark our framework on 8 ML datasets from Kaggle Competitions, which include 7 classification and 1 regression task, as well as 1 task being multi-target. Each competition contains a train and test split, a sample submission, and a description downloaded from the corresponding Kaggle page.
The dataset is split into train and validation sets within the pipeline with an 8:2 ratio.

Competitions are split evenly between classic (up to 2024) and modern (after 2024), coinciding with the knowledge cutoff in GPT-4o and GPT-4o-mini, which we adopt as LLMs. 

The normalized performance score, also used in AutoKaggle~\cite{li2024autokagglemultiagentframeworkautonomous}, is adopted as a quality measure:
\begin{equation}
\mathrm{NPS}= \begin{cases}\frac{1}{1+s}, & \text { if } s \text { is smaller the better } \\ s, & \text { otherwise }\end{cases}
\end{equation}
We directly include the normalized results from the aforementioned work, including the performance of the AutoKaggle and AIDE frameworks with GPT-4o.

\subsection{Results}

The experimental evaluation results for LightAutoDS-Tab are provided in Table~\ref{tab_res}.

\begin{table*}[t]
\caption{Results of experiments for LightAutoDS tools and existing solutions compared with human-designed results. The normalized performance score is used as a quality measure for solutions.}
\label{tab_res}
\setlength{\tabcolsep}{9.5pt}
\begin{tabular}{|c|ccc|cc|ccc|}
\hline
\multirow{2}{*}{Dataset} & \multicolumn{3}{c|}{Tools of LightAutoDS-Tab}                                                                                                                                                             & \multicolumn{2}{c|}{\begin{tabular}[c]{@{}c@{}}Existing\\ solutions\end{tabular}}           & \multicolumn{3}{c|}{\begin{tabular}[c]{@{}c@{}}Human results\\ from leaderboards\end{tabular}} \\ \cline{2-9} 
                         & \multicolumn{1}{c|}{\begin{tabular}[c]{@{}c@{}}LAMA\\ +LLM\end{tabular}} & \multicolumn{1}{c|}{\begin{tabular}[c]{@{}c@{}}Code\\ Gen\end{tabular}} & \begin{tabular}[c]{@{}c@{}}FEDOT\\ +LLM\end{tabular} & \multicolumn{1}{c|}{\begin{tabular}[c]{@{}c@{}}Auto\\ Kaggle\end{tabular}} & AIDE           & \multicolumn{1}{c|}{Q25}              & \multicolumn{1}{c|}{Q50}             & Q75             \\ \hline
Titanic                  & \multicolumn{1}{c|}{0.745}                                      & \multicolumn{1}{c|}{0.766}                                              & \textbf{0.780}                                       & \multicolumn{1}{c|}{0.767}                                        & 0.744          & \multicolumn{1}{c|}{\textit{0.766}}   & \multicolumn{1}{c|}{\textit{0.775}}  & \textit{0.778}  \\ \hline
Sp. Titanic              & \multicolumn{1}{c|}{\textbf{0.798}}                                      & \multicolumn{1}{c|}{0.788}                                              & 0.790                                                & \multicolumn{1}{c|}{0.771}                                                 & 0.793           & \multicolumn{1}{c|}{\textit{0.775}}   & \multicolumn{1}{c|}{\textit{0.792}}  & \textit{0.800}  \\ \hline
House Prices             & \multicolumn{1}{c|}{\textbf{0.886}}                                      & \multicolumn{1}{c|}{0.871}                                              & 0.882                                                & \multicolumn{1}{c|}{0.862}                                                 & 0.883          & \multicolumn{1}{c|}{\textit{0.857}}   & \multicolumn{1}{c|}{\textit{0.874}}  & \textit{0.884}  \\ \hline
Monsters                 & \multicolumn{1}{c|}{\textbf{0.774}}                                      & \multicolumn{1}{c|}{\textbf{0.774}}                                     & 0.733                                                & \multicolumn{1}{c|}{0.723}                                                 & 0.721          & \multicolumn{1}{c|}{\textit{0.726}}   & \multicolumn{1}{c|}{\textit{0.739}}  & \textit{0.747}  \\ \hline
Ac. Success              & \multicolumn{1}{c|}{\textbf{0.836}}                                      & \multicolumn{1}{c|}{0.828}                                              & 0.833                                                & \multicolumn{1}{c|}{0.820}                                                 & 0.835 & \multicolumn{1}{c|}{\textit{0.828}}   & \multicolumn{1}{c|}{\textit{0.834}}  & \textit{0.836}  \\ \hline
Bank Churm               & \multicolumn{1}{c|}{0.883}                                               & \multicolumn{1}{c|}{\textbf{0.885}}                                     & 0.881                                                & \multicolumn{1}{c|}{0.856}                                                 & 0.786           & \multicolumn{1}{c|}{\textit{0.870}}   & \multicolumn{1}{c|}{\textit{0.888}}  & \textit{0.891}  \\ \hline
Ob. Risk                 & \multicolumn{1}{c|}{\textbf{0.905}}                                      & \multicolumn{1}{c|}{0.888}                                              & 0.904                                                & \multicolumn{1}{c|}{0.896}                                                 & 0.896          & \multicolumn{1}{c|}{\textit{0.893}}   & \multicolumn{1}{c|}{\textit{0.903}}  & \textit{0.907}  \\ \hline
Plate Defect             & \multicolumn{1}{c|}{\textbf{0.886}}                                      & \multicolumn{1}{c|}{0.878}                                              & 0.883                                                & \multicolumn{1}{c|}{0.823}                                                 & -              & \multicolumn{1}{c|}{\textit{0.852}}   & \multicolumn{1}{c|}{\textit{0.879}}  & \textit{0.886}  \\ \hline \hline
Avg score                  & \multicolumn{1}{c|}{\textbf{0.839}}                                      & \multicolumn{1}{c|}{0.835}                                              & 0.835                                                & \multicolumn{1}{c|}{0.816}                                                 & 0.703          & \multicolumn{1}{c|}{0.82}             & \multicolumn{1}{c|}{0.836}           & 0.841           \\ \hline
\end{tabular}
\end{table*}

The results demonstrate that LightAutoDS-Tab achieves superior performance compared to AutoKaggle and AIDE. Furthermore, the comparative analysis of different tools within LightAutoDS-Tab confirms the validity of the proposed multi-AutoML implementation, as no single tool consistently delivers the best results across all cases.

The influence of LLM selection on CodeGen performance is analyzed in Table~\ref{tab_llms}, comparing GPT-4o and GigaChat2Max \cite{valentin2025gigachat}. The results demonstrate that the CodeGen output is highly sensitive to the choice of the underlying LLM.

\begin{table}[ht]
\setlength{\tabcolsep}{9.5pt}
\caption{ Ablation results of comparison of different LLMs used as CodeGen tool. The normalized performance score is used as a quality measure.}
\label{tab_llms}
\begin{tabular}{|c|cc|}
\hline
                          & \multicolumn{2}{c|}{LLM for CodeGen}                                \\ \cline{2-3} 
\multirow{-2}{*}{Dataset} & \multicolumn{1}{c|}{GPT4o}                         & GigaChat2Max   \\ \hline
Titanic                   & \multicolumn{1}{c|}{0.756}                         & \textbf{0.766} \\ \hline
Sp. Titanic               & \multicolumn{1}{c|}{\textbf{0.790}}                & 0.788          \\ \hline
House Prices              & \multicolumn{1}{c|}{0.868}                         & \textbf{0.871} \\ \hline
Monsters                  & \multicolumn{1}{c|}{0.755}                         & \textbf{0.774} \\ \hline
Ac. Success               & \multicolumn{1}{c|}{0.827}                         & \textbf{0.828} \\ \hline
Bank Churm                & \multicolumn{1}{c|}{0.884} & \textbf{0.885} \\ \hline
Ob. Risk                  & \multicolumn{1}{c|}{\textbf{0.899}}                & 0.888          \\ \hline
Plate Defect              & \multicolumn{1}{c|}{0.871}                         & \textbf{0.878} \\ \hline\hline
Avg score                       & \multicolumn{1}{c|}{0.832}                         & \textbf{0.835} \\ \hline
\end{tabular}
\end{table}

\section{Conclusion}
This paper introduces LightAutoDS-Tab (Figure~\ref{fig_schema}), a multi-AutoML agentic system designed to enhance the productivity of data scientists working with tabular data. LightAutoDS-Tab significantly increases automation by eliminating the need for manual coding while maintaining full interpretability throughout the end-to-end ML pipeline creation process. The framework drastically reduces development time and effort by allowing users to provide the dataset and a natural language query as input. Experimental results (Table~\ref{tab_res}) demonstrate that our approach outperforms other state-of-the-art open-source solutions on various data science tasks from Kaggle, indicating its ability to build high-quality, efficient models. Finally, we outline two directions for future research, including extending LightAutoDS-Tab to handle all types of structured data and integrating a more intelligent EDA system. 
The demonstration video is available in \url{https://www.youtube.com/watch?v=5e8eADd_HWE}.

\section{Limitations and Future Work}
Although significant progress has been made in developing LightAutoDS-Tab, our work has also identified two promising directions for future research. These represent key challenges and opportunities that have the potential to enhance the capabilities of automated data science systems significantly.

The first direction involves extending LightAutoDS-Tab to handle all types of structured data, including time series and sequential data, thereby enabling its application in domains such as time series forecasting and recommendation systems. This would require developing a new approach that allows each tool to be easily integrated into the framework.

The second direction focuses on integrating a more intelligent EDA system to minimize the risk of building models that suffer from data leakage or other anomalies that can negatively impact model performance.


\section{Acknowledgements}

This work is supported by the Ministry of Economic Development of the Russian Federation (IGK 000000C313925P4C0002), agreement No. 139-15-2025-010.

\bibliography{ref}

\newpage
\onecolumn
\appendix
\section[\appendixname~\thesection]{Technical appendix}
\label{app_a}

The examples of prompts for some agents are provided below. All prompts are available in the repository.

\subsection{FEDOT-specific prompt}
\label{app_cg_prompt}

\begin{lstlisting}
You are a helpful and intelligent assistant specializing in solving machine learning tasks. Your role is to complete and optimize Python scripts based on user instructions while adhering strictly to the specified constraints.
# About the Dataset:
[Task]
{reflection}

[Path to Dataset]
{dataset_path}

Below is the Python script you need to complete. Your implementation must begin with a Python code block (` ```python `) and strictly produce executable code without requiring further modifications.

[solution.py] 
```python
{skeleton}
```

**Key Rules and Constraints**:
1. Do **not delete any comments** in the provided code.
2. Do **not modify code enclosed between** the designated markers (`### comment ### code ### comment ###`). This code is autogenerated and will be regenerated upon project restarts.
3. You are **prohibited from using any methods or attributes** from the Fedot framework classes (e.g., `Fedot`, `Pipeline`), except those that are **already used in the provided code** or **explicitly mentioned in the comments**.
4. You are allowed to write and modify code **only within the 'USER CODE' sections**. All other sections will be regenerated upon project restarts.

Write the whole code below.
```python
\end{lstlisting}

\subsection{LightAutoML-specific prompts}
\begin{lstlisting}
You are an experienced machine learning developer who can formulate tasks in machine learning terms.
Your task is to create a configuration for training a machine learning model based on the input data.
For a regression task, use the metric (task_metric) "r2-score" and the task type (task_type) "reg".
For a classification task, use the metric (task_metric) "auc" and the task type (task_type) "binary".

Always respond only in the JSON format:
```
"task_type": "",
"target": "",
"task_metric": ""
```

Based on the user's task, column names, several rows from the dataset, and the file name, create a configuration for training.
User's task: {task}
File name: {file_name}
Column names: {df_columns}
Several rows from the dataset:
{df_head}
...

\end{lstlisting}

\subsection{AutoML router prompt}
\begin{lstlisting}
You are an experienced machine learning developer who understands the specific methods needed to solve a task.
Determine whether the user wants to solve the given task using the LightAutoML library, Fedot, or another automated machine learning method.
Carefully analyze the user's request to understand if automl, LightAutoML or Fedot is explicitly mentioned. If the request is general and does not explicitly mention automl, LightAutoML or Fedot, assume that the user does not want to use them and respond with the single word "NO".
If LightAutoML is specified, respond with the single word "LAMA".
If Fedot is specified, respond with the single word "FEDOT".
If automl is specified, respond with the single word "LAMA" or "FEDOT", you can choose.
\end{lstlisting}

\subsection{Fix solution prompt}
\label{app_fix_prompt}

\begin{lstlisting}
You are a senior machine learning engineer. Analyze the following information: the task description with reflections, the path to the dataset, the Python code from a previous solution, and the resulting stdout and stderr. Identify and correct the specific error that caused the failure without altering any other logic. Provide the complete corrected Python script in a code block.

# Task with Reflections
{reflection}

# Dataset Path
{dataset_path}

# Previous Python Solution
```python
{code_recent_solution}
```
{"# Execution Message: " + msg if msg else ""}

# Execution Output
```text
{stdout}
```

# Error Trace
```text
{stderr}
```

Write the full fixed code below.
```python
\end{lstlisting}

\subsection{Generate configuration prompt}
\label{app_gcprompt}

\begin{lstlisting}
You are a machine learning expert tasked with solving a given machine learning problem.  
Review the problem description provided within the `<problem-description>` section, including any reflections or additional context.  
Your objective is to define the optimal parameters for an automated machine learning model fitting framework, ensuring alignment with the stated goals, rules, and constraints of the task.  
If specific parameter values or constraints are not provided, use default values that are commonly accepted as best practices.

<problem-description>
{reflection}
</problem-description>
\end{lstlisting}

\subsection{Problem reflection prompt}
\label{app_prprompt}

\begin{lstlisting}
Please conduct a comprehensive analysis of the competition, focusing on the following aspects:
1. Competition Overview: Understand the background and context of the topic.
2. Files: Analyze each provided file, detailing its purpose and how it should be used in the competition.
3. Problem Definition: Clarify the problem's definition and requirements.
4. Data Information: Gather detailed information about the data, including its structure and contents.
    4.1 Data type:
        4.1.1. ID type: features that are unique identifiers for each data point, which will NOT be used in the model training.
        4.1.2. Numerical type: features that are numerical values.
        4.1.3. Categorical type: features that are categorical values.
        4.1.4 Datetime type: features that are datetime values.
    4.2 Detailed data description
5. Target Variable: Identify the target variable that needs to be predicted or optimized, which is provided in the training set but not in the test set.
6. Evaluation Metrics: Determine the evaluation metrics that will be used to assess the submissions.
7. Submission Format: Understand the required format for the final submission.
8. Other Key Aspects: Highlight any other important aspects that could influence the approach to the competition.
Ensure that the analysis is thorough, with a strong emphasis on :
1. Understanding the purpose and usage of each file provided.
2. Figuring out the target variable and evaluation metrics.
3. Classification of the features.

# Available Data File And Content in The File
{data_files_and_content}

# EDA
{dataset_eda}
\end{lstlisting}

\subsection{Reporter prompt}
\label{app_repprompt}

\begin{lstlisting}
You are an expert in machine learning tasked with evaluating and reporting on an ML model designed to address the problem.

Your report should adhere to the following instructions:  
- Be concise and styled like a Substack blog summary.  
- Structure the content using bullet points for readability.  
- Use Markdown formatting (e.g., headers, **bold text**, `code snippets`, tables) for clarity.  
- Explain key points in popular science language, accessible to readers from diverse fields.  
- Include code snippets and interpretation of results in layman-friendly terms.  
- Provide essential context while avoiding discussions of empty or missing values.  
- Do not include suggestions for next steps or future work.

**Report Outline:**  
1. **Overview**  
   - Problem description
   - Goal: Summarize the purpose of the model in plain terms for a general audience.

2. **Data Preprocessing**  
   - Describe the data preprocessing steps used before modeling in plain, accessible language.  
   - Provide illustrative examples to clarify specific preprocessing steps.  
    - If data normalization was applied, describe it as:  
    "Normalization ensures all features are on the same scale, improving model performance. For example, a scaling process converts values like 'age' (5-90) to 0-1."  
    - If missing values were imputed, describe it as:  
    "Missing values in the dataset were replaced using mean imputation to ensure uniformity. For example, in the column 'Income', the mean value of $50,000 was substituted for missing entries."

3. **Pipeline Summary**  
   - Summarize the steps in `{pipeline}` using accessible language and optionally include illustrative examples. 
   - Key Parameters:
    | Model   | Parameters   | Explanation   |
    |---------|--------------|---------------|
    | CatBoost| num_trees: 3000, learning_rate: 0.03, max_depth: 5, l2_leaf_reg: 0.01 | CatBoost was choosen because |
    | Model 2 | Parameter 2  | Explanation 2 |

4. **Code Highlights:**  
   - Include relevant code snippets wrapped in Markdown Python blocks:  
   ```python
   {code}
   ```
   - Add a brief explanation of what the code does and why it's a key.
   1. Data Preprocessing (Short Key Snippets)
   2. Model Training, Evaluation, Prediction
   3. Submission File Creation
   4. Other Key Snippets

5. **Metrics**  
   - Share performance metric `{metrics}` and briefly describe what each metric signifies (e.g., "Accuracy tells us how often the model gets it right"):

6. **Takeaways**  
   - Wrap up with a concise summary of results, emphasizing their significance in a real-world context. For example: "This model predicts X with an accuracy of Y%, demonstrating its potential in Z applications."

Engage your audience with a relatable, professional tone that simplifies complex ideas without oversimplifying the context. Ensure the report can resonate with both experts and non-experts alike.
\end{lstlisting}

\end{document}